\pdfoutput=1
\documentclass[11pt]{article}
\usepackage[final]{acl}
\usepackage{times}
\usepackage{latexsym}
\usepackage[T1]{fontenc}
\usepackage[utf8]{inputenc}
\usepackage{microtype}
\usepackage{graphicx}
\usepackage{booktabs}
\usepackage{array}
\usepackage{url}
\usepackage{enumitem}
\usepackage{hyperref}
\usepackage{makecell}
\usepackage{multirow}
\usepackage{amsmath}
\usepackage{amssymb}
\usepackage{float}
\usepackage{CJKutf8}
\newcommand{\zh}[1]{\begin{CJK}{UTF8}{gbsn}#1\end{CJK}}
\newcommand{\cjkbf}[1]{{\CJKfamily{gkai}#1}}

\usepackage{etoolbox}
\usepackage{subcaption}
\usepackage{placeins}
\usepackage{dblfloatfix}
\makeatletter
\ifacl@finalcopy
  \providecommand{\nolinenumbers}{}
\else
\AfterEndEnvironment{figure}{\linenumbers}
\AfterEndEnvironment{figure*}{\linenumbers}
\AfterEndEnvironment{table}{\linenumbers}
\AfterEndEnvironment{table*}{\linenumbers}
\fi
\makeatother

\title{Encoded but Not Decoded:\\
Layer-Localized Evidence for a Three-Level Gap in LLM Syntax}

\author{
  Zhenyan Lu \quad He Wang \quad Xiaohui Huang\Thanks{\bfseries\hspace{4.5pt}Corresponding author.} \\
  College of International Studies, \\
  National University of Defense Technology, Nanjing, China \\
  \texttt{\{lzy\_25, wanghe24\}@nudt.edu.cn} \\
  \texttt{huangxia@mail.ustc.edu.cn}
}

\begin{document}
\maketitle

\begin{abstract}
A language model can fail a syntactic test in two distinct ways: by not encoding the relevant structure, or by encoding it but failing to use it at the output. Behavioral evaluation alone cannot tell these apart. We propose a three-level evaluation framework (behavioral deployment, LM-head readout, and probe recoverability) measured on the same items under the same binary decision. Using a compact trilingual (English, Chinese, German) control-dependency benchmark, we find that probe recoverability exceeds or equals LM-head readout, which in turn exceeds or equals behavioral deployment, across seven models and all three languages in the aggregate. The recoverability surplus is never negative across all 14 (model, task) conditions. The disconnect concentrates in subject-control, where a nearest-noun heuristic gives the wrong answer. The single largest gap ($0.653$) appears on Qwen3-0.6B Instruct in question answering. The gap persists at Qwen3-14B Instruct. Instruction tuning degrades deployment more than encoding in percentage terms. We rule out option-position bias, late-layer erasure, output-formatting artifacts, and probe-training variance. The pattern is consistent with decoding that favors surface shortcuts, and the behavior--probe gap measures the strength of that preference. Activation patching shows the gap is layer-localized. Under instruction tuning, the LM-head-decoded layer shifts approximately ten layers later than the probe-decoded layer. These findings argue that behavioral evaluation understates what models encode, while probing alone overstates what they deploy.

\end{abstract}

\section{Introduction}
A syntactic error by a language model is ambiguous between two accounts. The structure may never have been encoded, or it was encoded and remained recoverable in the hidden states yet failed to reach the final output. Behavioral evaluation alone cannot distinguish them. Telling them apart requires measuring what the model encodes, not only what it outputs. The information is in the representation. The gap opens at the output.

Control dependencies provide a natural testbed for this distinction. In \emph{John told Mary to leave}, the understood subject of \emph{leave} is Mary (object control); in \emph{John promised Mary to leave}, it is John (subject control). Object-control aligns with a nearest-noun heuristic; subject-control overrides it, because the matrix subject is further from the lower predicate. Decoding by linear proximity therefore fails subject-control while answering object-control correctly.

To measure the encoding-deployment distinction on the same items, we propose a three-level framework. \textbf{Behavioral deployment} scores the correct continuation under a task-facing prompt \citep{linzen-etal-2016-assessing,warstadt-etal-2020-blimp}. \textbf{LM-head readout} projects intermediate hidden states through the model's own output map (the LM head) and scores candidate continuations layer by layer \citep{nostalgebraist-2020-logit,geva-etal-2021-transformer}. \textbf{Probe recoverability} reports the best accuracy a lightweight linear classifier achieves from any layer, as a ceiling on what a linear probe can recover \citep{belinkov-2022-probing,hewitt-liang-2019-designing}. The three measurements share the same 48 hand-curated minimal pairs under the same binary decision. The levels form a logical order: behavioral deployment is the tightest floor, probe recoverability is the highest ceiling, and LM-head readout sits between them, anchored in the model's own output geometry. That probing and behavior can disagree is not a hypothesis. Prior work has documented it empirically \citep{agarwal-etal-2025-mechanisms,he-etal-2025-large-language,waldis-etal-2024-holmes}. We ask whether the disagreement has a localizable structure consistent with a decoder that reaches for surface heuristics at the output when richer information is recoverable internally \citep{mccoy-etal-2019-right,geirhos-etal-2020-shortcut}.

Our benchmark is 48 hand-curated minimal-pair items in English, Chinese, and German \citep{de-marneffe-etal-2021-universal,sanches-duran-etal-2025-extending}, each in object-control and subject-control versions under question-answering and paraphrase-selection formats. The main model family is Qwen3 (0.6B base, 0.6B instruct, 14B instruct); Llama-3.2 and Gemma-4 serve as external-family checks. The three-level ordering behavior $\leq$ LM-head $\leq$ probe holds across all seven models and all three languages in the aggregate (Section~\ref{sec:results}). The effect concentrates in subject-control: the sharpest case is Qwen3-0.6B Instruct QA subject-control, behavior 0.250 against probe 0.903 (Figure~\ref{fig:framework}). Scaling to 14B does not close the subject-control gap, and instruction tuning degrades deployment more than encoding in percentage terms. A counterbalanced-ordering ablation rules out option-position artifacts \citep{shen-etal-2025-revisiting}.

We pose three research questions, each paired with a testable hypothesis:
\begin{itemize}
    \item \textbf{RQ1 (consistency).} Does the ordering behavior $\leq$ LM-head readout $\leq$ probe hold across models, languages, and control types? \\ \textbf{H1.} The ordering holds in the aggregate, and the recoverability surplus concentrates in subject-control.
    \item \textbf{RQ2 (localization).} Where does the subject-control deficit live: in the encoding, in the output geometry, or in deployment? \\ \textbf{H2.} The deficit is a deployment deficit: probe recoverability stays high where behavioral accuracy collapses.
    \item \textbf{RQ3 (locus of change).} Do scaling and instruction tuning shift the gap, and at which level? \\ \textbf{H3.} Scaling to 14B closes only the object-control gap; instruction tuning degrades deployment more than encoding in percentage terms and shifts the causally relevant readout layer later.
\end{itemize}

Our three-level framework triangulates the behavior--probe disconnect and localizes it at the representation-to-output interface. In syntactic control, the evidence is consistent with shortcut-preferring decoding, concentrated in subject-control, resistant to scaling, and sharpened by instruction tuning. Activation patching (Appendix~\ref{sec:appendix_steering}) gives preliminary causal evidence that in instruction-tuned models the LM-head-decoded layer shifts approximately ten layers later than the probe-decoded layer. Our benchmark and ablation are reusable components; code and items are available at \href{https://github.com/camel-luv/encoded-but-not-decoded}{https://github.com/camel-luv/encoded-but-not-decoded}.

\section{Related Work}
Two evaluation traditions for syntactic knowledge in language models have developed largely in parallel. Behavioral evaluation compares the probabilities a model assigns to matched grammatical and ungrammatical strings \citep{linzen-etal-2016-assessing,gulordava-etal-2018-colorless,warstadt-etal-2020-blimp,gauthier-etal-2020-syntaxgym}. Representational probing asks what a lightweight classifier can recover from hidden states independently of behavior \citep{hewitt-manning-2019-structural,tenney-etal-2019-what,tenney-etal-2019-bert}, and has recently been extended to LLMs to test whether hierarchical syntactic structure (e.g., the control--raising distinction) is encoded \citep{kennedy-2025-evidence}. Each tradition is methodologically silent on the other's measurement. Behavioral benchmarks cannot access representations, and probing classifiers cannot validate output use. The three-level framework we propose measures both on the same items and inserts LM-head readout between them.

The two traditions diverge empirically. \citet{agarwal-etal-2025-mechanisms} show that syntactic probes do not reliably predict targeted syntactic evaluation outcomes; \citet{he-etal-2025-large-language} document a performance-competence distinction in which hidden states carry more information than final outputs express. \citet{zhu-etal-2025-llm} report a parallel effect. Internal representations encode question difficulty that behavior does not exhaust. Probing work on grammatical number makes the ``recoverability-versus-usage'' separation explicit within syntax. What a probe can recover from representations and what the model uses at the output are distinct quantities \citep{lasri-etal-2022-usage}. The Holmes benchmark extends this pattern across a broad suite of linguistic phenomena and architectures \citep{waldis-etal-2024-holmes}. These studies document the disconnect without localizing where in the pipeline it opens or whether it varies with linguistic subtype; our three-level framework is that localization tool.

Probing has a known methodological vulnerability. A linear probe trained on top of frozen representations can succeed for reasons that reflect the probe's own capacity rather than the model's internal computation \citep{belinkov-2022-probing,hewitt-liang-2019-designing}, and probing accuracy alone does not establish that the model uses the recovered information \citep{ravichander-etal-2021-probing}. LM-head readout addresses both limitations. The logit-lens tradition has established that intermediate hidden states projected through the model's own output embedding carry interpretable layer-wise predictions \citep{nostalgebraist-2020-logit,geva-etal-2021-transformer,geva-etal-2022-vocabulary}. We use it to anchor the three-level comparison at the output interface. Prompt- and label-sensitivity of probing \citep{shen-etal-2025-revisiting} is handled by a counterbalanced-ordering ablation; causal-mediation analysis, causal abstraction, and activation editing replace correlational probing with intervention \citep{vig-etal-2020-causal,geiger-etal-2021-causal,meng-etal-2022-locating}.

Control dependencies are the linguistic test case that makes this triangulation interpretable. Control is a syntax-semantics interface phenomenon. The controller of the embedded predicate is determined by the thematic structure of the matrix verb, not by linear adjacency, with both movement-based \citep{hornstein-1999-movement,boeckx-hornstein-2007-non-obligatory} and PRO-based \citep{landau-2013-control} accounts in generative grammar. The object/subject contrast makes control productive as a diagnostic. Object-control aligns with linear proximity (the controller is the immediately preverbal noun), while subject-control overrides it. Corpus evidence shows that such surface heuristics describe much of controller resolution in natural text \citep{stengel-eskin-van-durme-2022-curious}, which is why subject-control is the sharp place to test them. Earlier work has observed the object/subject asymmetry computationally \citep{de-dios-flores-etal-2023-dependency}.

Three languages (English, Chinese, German) harden the empirical claim against single-language artifacts. Cross-lingual comparisons reflect more than typology. Tokenization, pretraining balance, and instruction tuning all contribute \citep{hu-etal-2025-quantifying,goworek-dubossarsky-2025-multilinguality,chirkova-nikoulina-2024-zero-shot}.

\section{Methods}
\label{sec:methods}
\subsection{Benchmark design}

We study control dependencies. The embedded predicate has an understood subject that is not overtly realized inside the embedded clause, and the benchmark separates object-control items, where the matrix object is the controller, from subject-control items, where the matrix subject controls (Appendix~\ref{sec:appendix_benchmark}). This contrast drives the paper's central result, because deployment failures concentrate in subject-control. The 48 hand-curated items split evenly across English, Chinese, and German (8 object/8 subject per language). Statistical leverage comes from minimal-pair structure rather than item count. Items within a language share their NPs and event semantics, differing only in the matrix verb's controller assignment, and cross-condition claims aggregate over models, languages, and layers.

 Each item carries two candidate controller nouns, the lower predicate, a gold controller annotation, a language label, and a control-type label. Behavior is evaluated under two task formats, question answering and paraphrase selection (Table~\ref{tab:task_format} in Appendix~\ref{sec:appendix_benchmark}), both pairwise log-probability scoring over two explicit continuations, so deployment failure is not tied to one template. For the focal Qwen models we additionally run a counterbalanced-ordering variant, averaging both candidate orderings so the result is not driven by option position (Section~\ref{sec:results}).

Item construction follows Universal Dependencies (UD) and Enhanced UD conventions \citep{de-marneffe-etal-2021-universal,sanches-duran-etal-2025-extending}. Each item instantiates a matrix predicate with an open-clausal complement (UD \texttt{xcomp}) whose embedded subject is propagated by EUD \texttt{nsubj:xsubj}. Object-control items lexicalize the controller as the matrix object, subject-control as the matrix subject. All matrix predicates are attested with \texttt{xcomp} in standard UD treebanks (English EWT, German HDT, Chinese GSDSimp).

\subsection{Three levels of access}\label{sec:levels}

For each benchmark item $x$ with candidate controllers $c_a, c_b$ and gold controller $c^*(x) \in \{c_a, c_b\}$, we define three accuracies on the identical binary forced choice between $c^*(x)$ and the distractor $\bar{c}(x)$. Differences between the three levels are interpretable as gaps in how much controller information is accessible at each interface, not as differences in task or label format.

\paragraph{Behavioral deployment.}
For prompt format $p \in \{\mathrm{QA}, \mathrm{Para}\}$, the behavioral score is $s_{\mathrm{B}}(c, x) = \log P_\theta(c \mid p(x))$, and behavioral accuracy is the fraction of items where the gold candidate outscores the distractor: $\mathrm{Acc_B}(p) = \Pr_{x \sim \mathcal{D}}[\, s_{\mathrm{B}}(c^*, x) > s_{\mathrm{B}}(\bar{c}, x) \,]$. This level reflects both what the model encodes and what its output interface can express.

\paragraph{LM-head readout.}
LM-head readout is a raw diagnostic. It applies the model's own output map to intermediate states without calibration, prompt, or training. Let $h_l(x) \in \mathbb{R}^d$ be the residual-stream hidden state at layer $l$ for item $x$ at the scoring position, and let $W_U \in \mathbb{R}^{|V| \times d}$ be the model's unembedding matrix (the LM head). The LM-head score is $s_{\mathrm{LM}}(c, x, l) = (W_U h_l(x))_c$, scored at the candidate's first sub-token, with $\mathrm{Acc_{LM}}(l) = \Pr_{x \sim \mathcal{D}}[\, s_{\mathrm{LM}}(c^*, x, l) > s_{\mathrm{LM}}(\bar{c}, x, l) \,]$. Because $W_U$ is the model's own output map, $\mathrm{Acc_{LM}}$ cannot be attributed to probe overfitting. We report the summary $\mathrm{Acc_{LM}^*} = \max_l \mathrm{Acc_{LM}}(l)$ and the layerwise trajectory $\{\mathrm{Acc_{LM}}(l)\}_l$, which later distinguishes late-layer erasure from readout-bottleneck accounts.

The interpretation of $\mathrm{Acc_{LM}}$ rests on an architectural precondition. $W_U$ is trained against the residual stream at the final layer, and intermediate layers face no training pressure to remain geometrically compatible with the unembedding map \citep{elhoushi-etal-2024-layerskip,yom-din-etal-2024-jump}. An intermediate-layer LM-head score can therefore reflect geometric incompatibility rather than absent controller information \citep{wei-etal-2025-adadecode,wendler-etal-2024-llamas}, and on its own it is not evidence of a deployment failure. The behavior--probe contrast is what carries that claim.

 This is the ``recoverability-versus-usage'' separation that probing analyses must keep explicit \citep{ravichander-etal-2021-probing,lasri-etal-2022-usage}. A probe ceiling measures recoverability, and an LM-head readout measures compatibility with the output map. The precondition is cleanest for tied-embedding models. Among the seven main models, Gemma-4 E4B is the only model whose architecture most plausibly trains intermediate layers to remain compatible with the unembedding map (Per-Layer Embeddings, Section~\ref{sec:rq1_crossfamily}).

\paragraph{Probe recoverability.}
Because the probe decides between two candidates, feature modes must include both candidates or one candidate plus the predicate: $m \in \{ab, av, bv, abv, v\}$ (Table~\ref{tab:appendix_probe_modes}). For each layer $l$ and feature mode $m$ (concatenations of the hidden states at the candidate-A, candidate-B, and lower-predicate token positions) we train a linear probe $g_{l,m}(h) = \sigma(w_{l,m}^{\top} h + b_{l,m})$, fit by leave-one-out cross-validation against the binary $c^*$ vs.\ $\bar{c}$ label. Each probe is a logistic regression with bias, trained by Adam (learning rate $10^{-2}$) for a fixed 100 epochs on binary cross-entropy, with no weight decay and no input normalization.

Probe accuracy $\mathrm{Acc_P}(l, m)$ is the corresponding leave-one-out accuracy, and the probe summary is $\mathrm{Acc_P^*} = \max_{l,\,m} \mathrm{Acc_P}(l, m)$. Each probe is trained with three independent random seeds (1729, 2718, 3141). The main table reports the seed-1729 run, and Table~\ref{tab:probe_seed_robustness} confirms probe ceilings are stable across seeds (max $\sigma = 0.064$, roughly three of 48 items). $\mathrm{Acc_P^*}$ is a ceiling estimate of what a linear decoder can extract from the most informative layer of the hidden-state stack, not a prediction of what the model will deploy at the output. Because $\mathrm{Acc_P^*}$ maximizes over layers and modes, the reported ceiling carries selection optimism. The verb-out cross-validation of Appendix~\ref{sec:appendix_lexical} bounds it directly.

Beyond raw accuracies we report three descriptive quantities, the \emph{recoverability surplus} ($\mathrm{Acc_P^*} - \overline{\mathrm{Acc_B}}$), the \emph{behavior-to-recoverability ratio} ($\overline{\mathrm{Acc_B}} / \mathrm{Acc_P^*}$), and the \emph{subject-control deficit} ($\overline{\mathrm{Acc_B}}^{\,\mathrm{obj}} - \overline{\mathrm{Acc_B}}^{\,\mathrm{subj}}$), where $\overline{\mathrm{Acc_B}}$ is mean behavior across QA and paraphrase. Full definitions and the rationale for mean over best behavior are in Appendix~\ref{sec:appendix_metrics}.

\subsection{Models}

The trilingual evaluation centers on three Qwen3 models (0.6B base, 0.6B instruct, 14B instruct). The base/instruct pair gives the clearest dissociation, and 14B tests whether the phenomenon survives scaling. External-family checks come from Llama-3.2 (1B base/instruct) as the primary replication target and Gemma-4 E4B (base and Instruct) as an architectural contrast on several axes at once (Section~\ref{sec:rq1_crossfamily}). Two larger base/instruct pairs (Qwen3-1.7B, Llama-3.2-3B) extend the scaling axis for the surplus-compression analysis (Section~\ref{sec:rq3_instruction}).

\begin{table*}[!t]
\centering
\caption{Main trilingual model comparison across behavioral deployment, LM-head readout, and probe recoverability.}
\label{tab:main_model_comparison}
\resizebox{\linewidth}{!}{%
\begin{tabular}{lccccccc}
\hline
Model & QA & Paraphrase & Mean Behavior & Best LM-head & Best Probe & Surplus & B/R Ratio \\
\hline
Qwen3-0.6B Base & 0.542 & 0.646 & 0.594 & 0.708 & 0.938 & 0.344 & 0.633 \\
Qwen3-0.6B Instruct & 0.417 & 0.604 & 0.510 & 0.667 & 0.833 & 0.323 & 0.612 \\
Qwen3-14B Instruct & 0.812 & 0.771 & 0.792 & 0.792 & 1.000 & 0.208 & 0.792 \\
Llama-3.2-1B Base & 0.521 & 0.625 & 0.573 & 0.688 & 0.812 & 0.240 & 0.705 \\
Llama-3.2-1B Instruct & 0.479 & 0.542 & 0.510 & 0.583 & 0.625 & 0.115 & 0.817 \\
Gemma-4 E4B & 0.521 & 0.604 & 0.562 & 0.604 & 0.604 & 0.042 & 0.931 \\
Gemma-4 E4B Instruct & 0.438 & 0.500 & 0.469 & 0.583 & 0.667 & 0.198 & 0.703 \\
\hline
\end{tabular}%
}
\end{table*}

\begin{figure*}[t]
    \centering
    \includegraphics[width=0.85\linewidth]{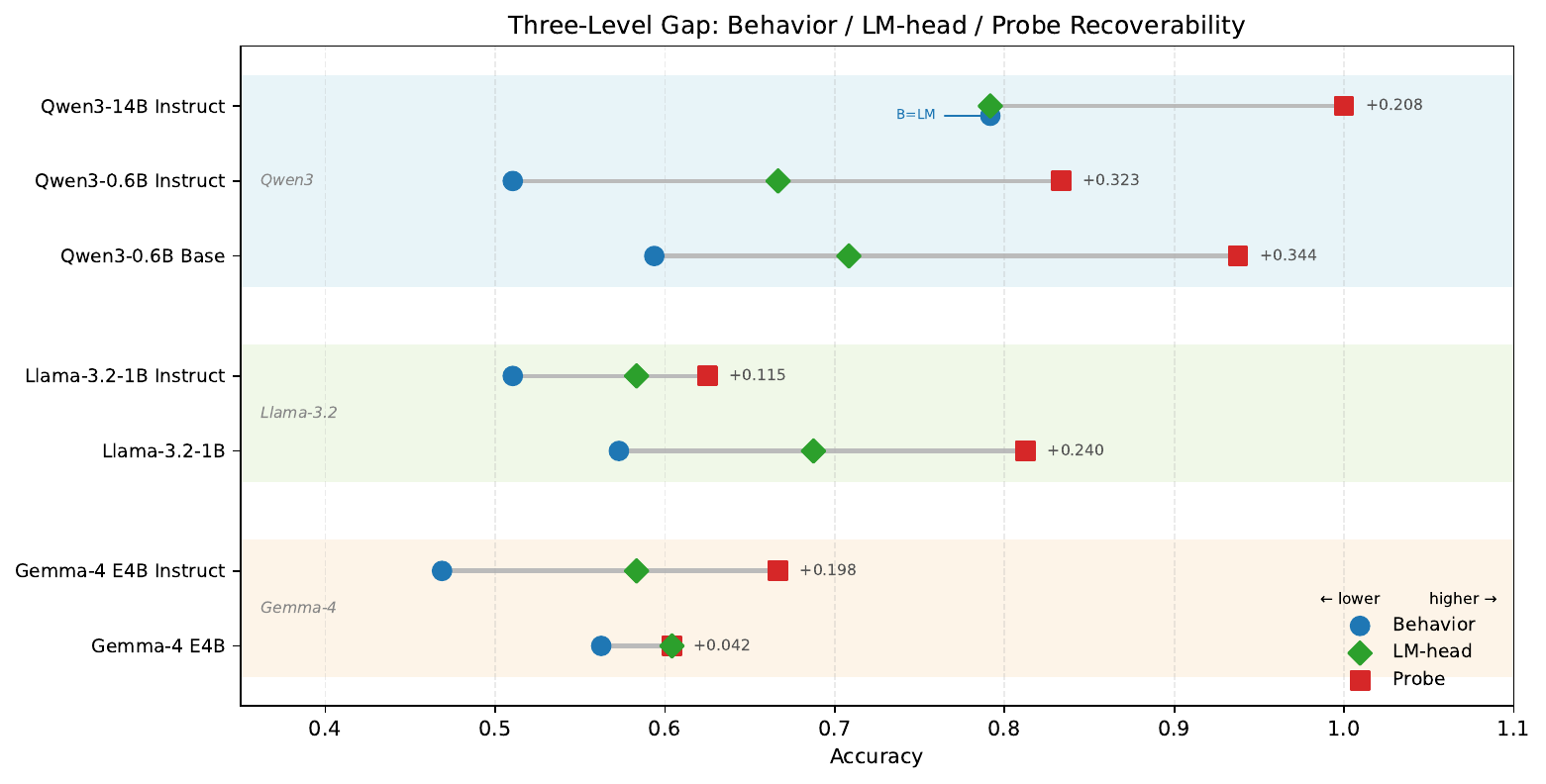}
    \caption{Three-level gap across all seven models (right: recoverability surplus).}
    \label{fig:model_gap}
\end{figure*}

\section{Results}
\label{sec:results}

\subsection{Consistency of the three-level ordering}
\label{sec:rq1_ordering}

\begin{figure}[t]
    \centering
    \includegraphics[width=0.9\linewidth]{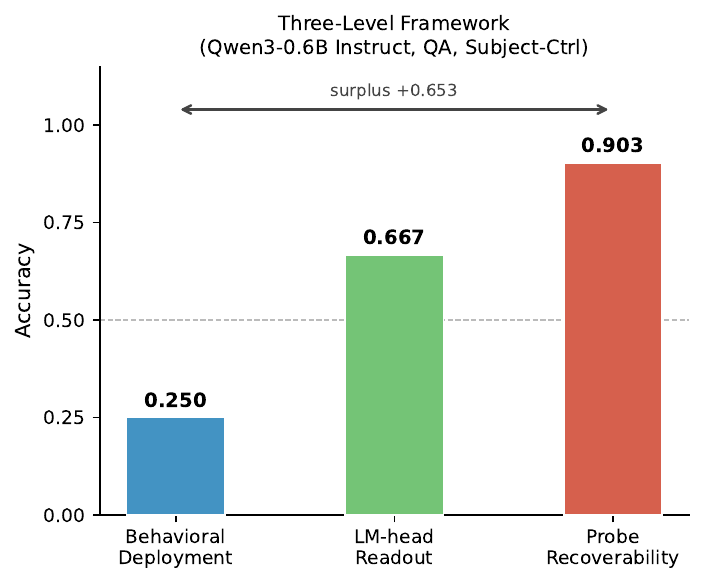}
    \caption{Three-level framework on the hardest case (Qwen3-0.6B Instruct, QA, subject-control), where behavioral deployment (0.250) falls far below LM-head readout (0.667), which falls far below probe recoverability (0.903).}
    \label{fig:framework}
\end{figure}

On Qwen3-0.6B Instruct subject-control QA, behavioral accuracy is 0.250 while probe accuracy reaches 0.903 (Figure~\ref{fig:framework}). The 0.653 behavior--probe gap is the largest in the experiment.

The broader pattern is not a quirk. Across all seven models and all three languages, probe recoverability $\geq$ LM-head readout $\geq$ mean behavioral accuracy (Table~\ref{tab:main_model_comparison}, Figure~\ref{fig:model_gap}); no model inverts. The recoverability surplus $\mathrm{Acc_P^*} - \overline{\mathrm{Acc_B}}$ averages $+0.210$ across 14 (model, task) conditions, and the ordering holds at every scale in each family and in every language (per-language breakdown in Table~\ref{tab:per_language_breakdown}). A regularity that survives this many axes is structural, not probe noise, nor does the LM-head merely re-read final behavior. Table~\ref{tab:main_model_comparison} is descriptive; the controls below address sampling and robustness.

The aggregate ordering admits per-language exceptions of bounded magnitude. Three of the 21 (model, language) cells invert, each within one item of per-language resolution ($1/16 \approx 0.062$). Only Gemma-4 E4B German exceeds that scale, at a 1.5-item swing (behavior $0.594$ vs.\ LM-head $0.500$). Every other inversion is at or below one item (Appendix~\ref{sec:appendix_perlang}). The ordering is a property of the aggregate, not of every cell. Precision at the 48-item scale is bounded. A cluster bootstrap (B = 10{,}000, clustered by the 16 translation groups) yields per-cell 95\% CIs of width 0.23--0.42, and leave-one-family-out sensitivity keeps overall behavioral accuracy within 0.549--0.552 (whether holding out Qwen3, Llama, or Gemma). This answers RQ1. The ordering holds in the aggregate with exceptions bounded by item-level resolution, and its subject-control concentration survives scaling and family variation.

\subsection{Cross-family replication}
\label{sec:rq1_crossfamily}

Llama-3.2 shows a complementary pattern. The 1B base yields $0.573 < 0.688 < 0.812$ (surplus $+0.240$). The 1B instruct narrows to $0.510 < 0.583 < 0.625$ (surplus $+0.115$). The mechanism differs from Qwen3's. Behavior drops $0.063$ ($-11.0\%$), LM-head readout drops $0.105$ ($-15.3\%$), and the probe ceiling drops $0.187$ ($-23.0\%$). Instruction tuning of Llama-3.2-1B degrades recoverability more than deployment, whereas in Qwen3-0.6B deployment degrades more than recoverability ($-14.1\%$ vs. $-11.2\%$). The shared outcome across families is a narrowed surplus and a persistent gap.

Gemma-4 E4B is an architectural boundary. Differing on multiple axes at once (multimodal pretraining, Per-Layer Embeddings, hybrid attention, smaller effective capacity (~4B)), its profile is $0.562 < 0.604 \approx 0.604$ (surplus only $+0.042$). LM-head and probe ceiling have converged, which we treat as architectural-axis dependence of the framework, not a counterexample. The three-seed probe ceilings are $0.569 \pm 0.064$ (base) and $0.660 \pm 0.052$ (instruct); across seeds, both ceilings remain far below the Qwen3 and Llama-3.2 ceilings and the boundary characterization holds. This completes RQ1. The ordering and its subject-control concentration are not artifacts of one pretraining family.

\subsection{Alternative explanations}
\label{sec:rq2_alternatives}

\begin{figure}[t]
    \centering
    \includegraphics[width=\linewidth]{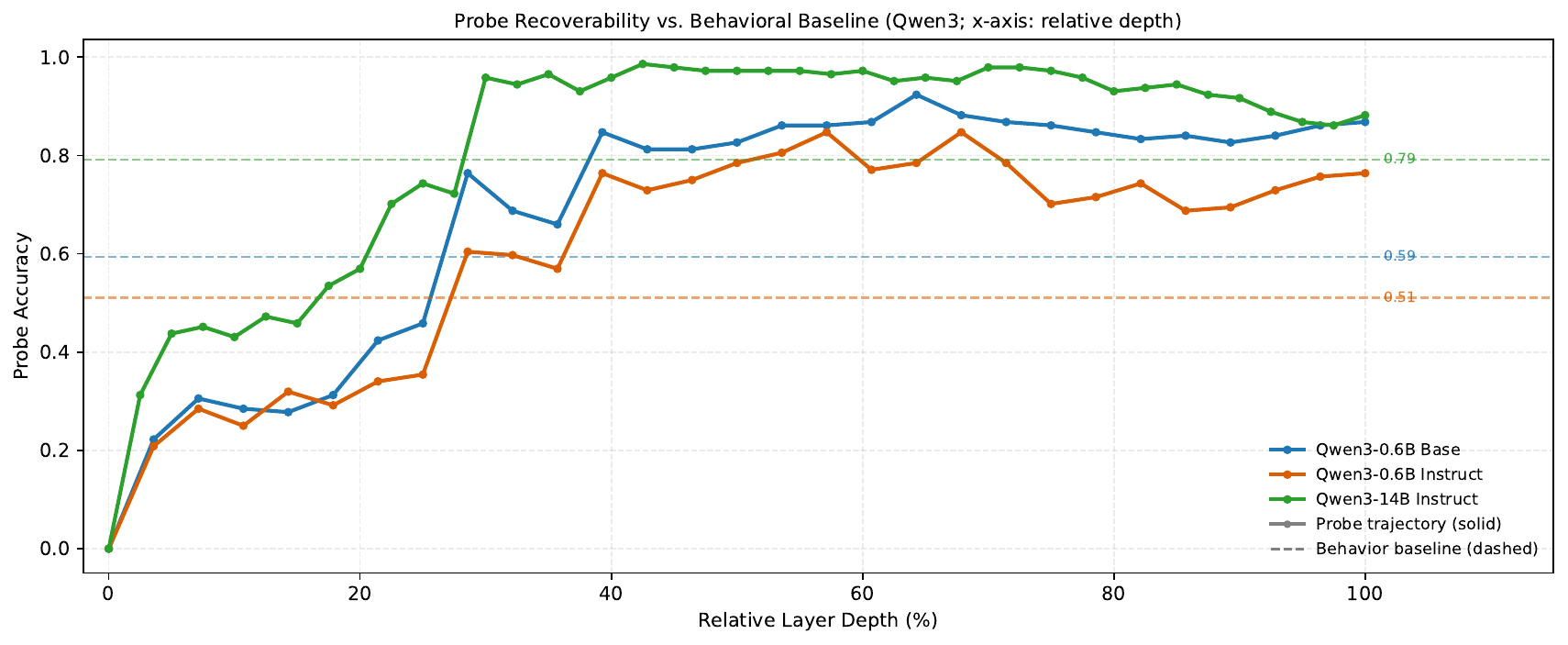}
    \caption{Probe accuracy by relative layer depth across Qwen3 models.}
    \label{fig:probe_traj}
\end{figure}

\textbf{Option-position bias.} The counterbalanced-ordering variant yields debiased forced-choice $0.500$ and label-choice $0.562$ (Table~\ref{tab:readout_cleaning}), both below LM-head $0.667$ and probe $0.833$.

\textbf{Late-layer erasure.} Probe accuracy stays high through final layers (Fig.~\ref{fig:probe_traj}).

\textbf{Output-formatting artifact.} With no prompt and no trained classifier, the gap persists at the unembedding map.

\textbf{Probe-training variance.} Three-seed retrain shows max $\sigma = 0.064$ (Gemma-4 E4B base), $\leq 0.052$ elsewhere (Tab.~\ref{tab:probe_seed_robustness}).

\textbf{Prompt-format robustness.} A task-free third prompt preserves the three-level ordering on five of seven models (e.g., Llama-3.2-1B Instruct behavior $0.719$, LM-head $0.812$). The two exceptions are Qwen3-0.6B base and Qwen3-14B Instruct, where the LM-head readout falls at or just below behavior.

None of the five explains the gap.
\subsection{Scaling effects}
\label{sec:rq2_subject14b}

If the deployment failure were a capacity limitation of small models, a strong 14B model should close the behavior--probe gap. It does for object-control. It does not for subject-control. On object-control paraphrase, Qwen3-14B Instruct reaches behavior 0.958 against probe 1.000 (gap 0.042, behavior at ceiling). On subject-control paraphrase, the same model reaches 0.583 against probe 1.000 (gap 0.417). The probe recovers the assignment perfectly from hidden states the model uses to answer with less-than-60\% accuracy. Qwen3-14B Instruct is not weaker in representation but in deployment, and only on the control type that contradicts linear adjacency.

The type-level breakdown across the focal Qwen models (Figure~\ref{fig:type_gap}; full numbers in Appendix~\ref{sec:appendix_typegap}) makes the asymmetry quantitative. The object-control gap shrinks with scale and approaches zero at 14B, while the subject-control gap does not shrink. This answers RQ2 and completes the scaling half of RQ3. Scaling raises object-control deployment to ceiling but leaves the subject-control deficit open.

\subsection{Instruction tuning and the deployment gap}
\label{sec:rq3_instruction}

\begin{figure}[t]
    \centering
    \includegraphics[width=\linewidth]{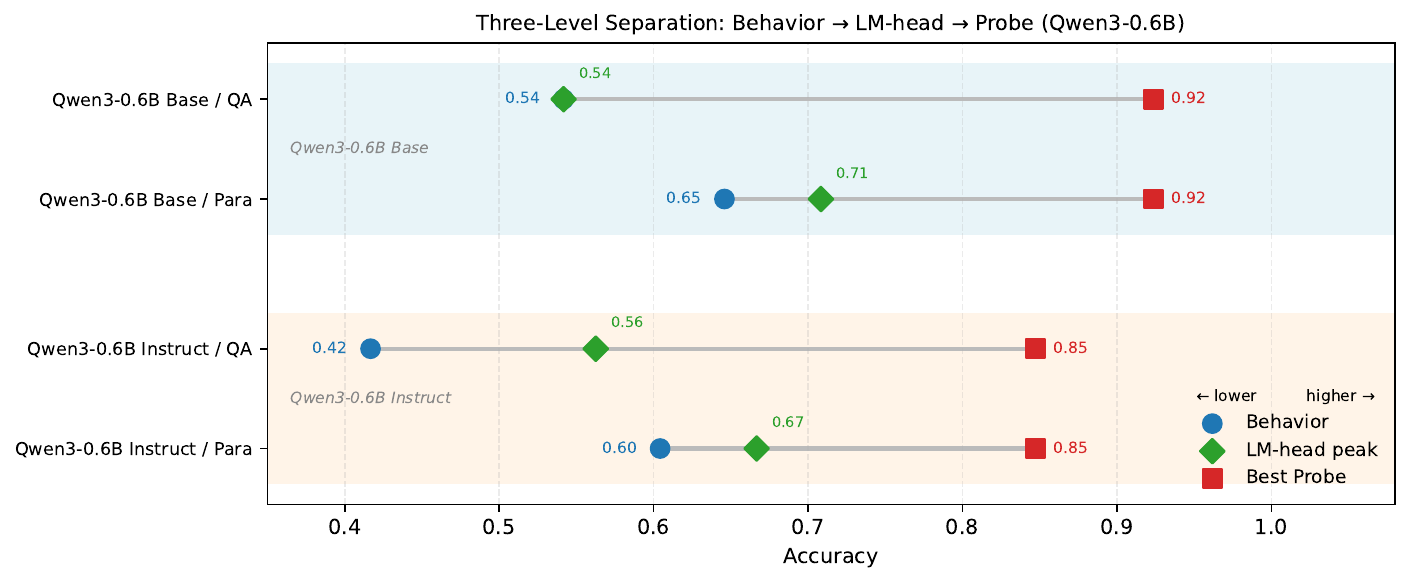}
    \caption{Three-level separation for Qwen3-0.6B base and instruct, by task, on the same items.}
    \label{fig:lmhead}
\end{figure}

The Qwen3-0.6B base/instruct pair shows the phenomenon most sharply because instruction tuning moves all three levels by different amounts (Figure~\ref{fig:lmhead}). From base to instruct, mean behavior drops $0.594 \rightarrow 0.510$ ($-0.084$, $-14.1\%$ relative), the probe ceiling drops $0.938 \rightarrow 0.833$ ($-0.105$, $-11.2\%$ relative), and LM-head readout moves $0.708 \rightarrow 0.667$. Both degrade, but deployment more than encoding in percentage terms (losses $-14.1\%$ vs.\ $-11.2\%$, surplus narrowing $0.344 \rightarrow 0.323$). Instruction tuning disproportionately affects deployment. The instructed model retains most of what it encoded and deploys less of it.

If instruction tuning had erased controller information, LM-head accuracy would collapse toward behavior. It does not. Instructed Qwen3-0.6B projects hidden states through its own output embedding and recovers the correct controller on $66.7\%$ of items, while task-facing behavior reaches only $51.0\%$ on the same items. The middle level separates ``less encoded'' from ``less deployed.'' Activation patching (Appendix~\ref{sec:appendix_steering}) sharpens this. In base models the layer where patching causally rescues controller assignment coincides with the probe-decoded layer, while in instruction-tuned models it shifts approximately ten layers later. This addresses the instruction-tuning half of RQ3. The gap shifts because deployment degrades more than encoding, not because encoding collapses.

The surplus compression amplifies with scale.
The Qwen3-0.6B pair narrows the surplus by $0.021$ under instruction tuning. Two larger base/instruct pairs run on the same 48 items under the same protocol (Qwen3-1.7B: probe ceiling $0.986 \rightarrow 0.958$, surplus $0.278 \rightarrow 0.188$, $\Delta -0.090$; Llama-3.2-3B: probe ceiling $0.958 \rightarrow 0.750$, surplus $0.250 \rightarrow 0.042$, $\Delta -0.208$) show that the compression deepens with scale and replicates beyond the Qwen3 family. The Llama-3.2-3B Instruct surplus collapses to $+0.042$, the same value that marks the Gemma-4 architectural boundary (\S\ref{sec:rq1_crossfamily}); the deployment--encoding distinction the framework draws at small scale is therefore not stationary, and where it disappears at still larger scale remains open. The mechanism differs across pairs. In Qwen3-1.7B, instruction tuning raises deployment ($0.708 \rightarrow 0.771$) yet the surplus still narrows because the probe ceiling drops in percentage terms. In Llama-3.2-3B, deployment stays flat ($0.708$) while the probe ceiling alone drops by $0.208$. All three pairs narrow the surplus. The harder-hit level differs across pairs.

\begin{figure}[t]
    \centering
    \includegraphics[width=0.85\linewidth]{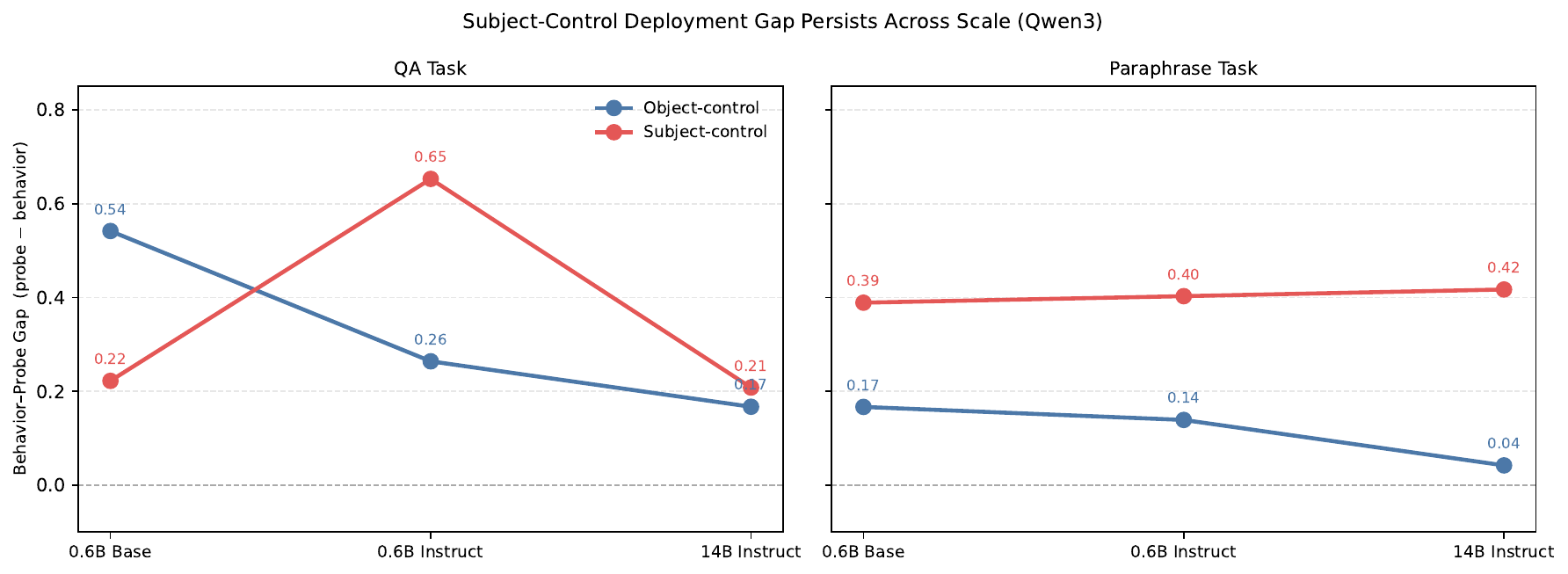}
    \caption{Subject-control deployment gap (probe $-$ behavior) across scale.}
    \label{fig:type_gap}
\end{figure}

\section{Discussion}
The three-level framework turns an empirical complaint into a localized diagnosis. Prior work has documented behavior--probe disagreement \citep{agarwal-etal-2025-mechanisms,he-etal-2025-large-language,waldis-etal-2024-holmes}. Our contribution is to show that inserting LM-head readout as an intermediate measurement gives the gap structure. The standard probing objection---that linear probes can extract information the model itself cannot use \citep{belinkov-2022-probing,hewitt-liang-2019-designing}---does not apply to LM-head readout, which uses the model's own output projection instead of a trained classifier \citep{nostalgebraist-2020-logit,geva-etal-2021-transformer}. At 14B, behavior and LM-head coincide at $0.792$ while the probe reaches $1.000$, so neither gap is an artifact of probe capacity or task format. Four observations point in the same direction. Ordering holds across seven models and three languages, deployment loss under instruction tuning is disproportionate, the failure concentrates in subject-control, and the deficit persists at 14B. The final decoding step selectively fails to use controller information the model both encodes and can project through its own output geometry.

\paragraph{Erasure and formatting as incomplete accounts.}
Late-layer erasure and output-format artifacts are the two cleanest alternative accounts, and Results~\S\ref{sec:rq2_alternatives} rules out either as a complete account. Probe recoverability stays high through the final layers, debiased behavioral accuracy stays well below the LM-head peak, and a task-free third prompt preserves the ordering on five of seven models. A mixed account remains plausible. Later representations become less reliably controller-aligned, and the final decoding step amplifies the residual mismatch rather than smoothing it. Neither process alone is sufficient.

\paragraph{Subject-control as the locus of deployment failure.}
Subject-control is the control type whose correct resolution actively contradicts linear proximity. A nearest-noun heuristic gets object-control right and subject-control wrong. This matches the shortcut-learning phenomenon, where models rely on surface heuristics at the output even when richer information is linearly recoverable from their representations \citep{mccoy-etal-2019-right,geirhos-etal-2020-shortcut}. Control thus provides a built-in contrast. Within one construction family, the heuristic succeeds on one subtype and fails on the other.

Three independent observations are consistent with this interpretation. At 14B, the subject-control probe ceiling is perfect while behavior on the same items remains far lower, so representational capacity is not the bottleneck. Instruction tuning degrades deployment more than encoding in percentage terms, so the degradation does not target the representation itself. The failure concentrates in the subtype where the heuristic is wrong. This is the pattern a decoder reaching for the shortcut would produce.

\paragraph{Two interpretations of the Gemma-4 E4B boundary.}
Gemma-4 E4B is the only model whose architecture most plausibly trains intermediate layers to remain compatible with the unembedding map (Per-Layer Embeddings, Section~\ref{sec:methods}). This gives its near-zero surplus ($+0.042$, LM-head $0.604 \approx$ probe $0.604$) two compatible interpretations. Under the boundary interpretation adopted in Section~\ref{sec:rq1_crossfamily}, the model differs on several architectural axes at once, the framework does not apply cleanly, and the near-zero gap carries no evidence either way. Under the stronger interpretation, Gemma-4 is the one model in which the raw-diagnostic precondition is closest to being satisfied, so its LM-head readout is the most trustworthy in the study, and the deficit is smallest exactly there. That pattern would support the geometric account of the gap. Our evidence does not separate the two interpretations, so we present both and rest the framework's cross-family claims on Qwen3 and Llama-3.2.

\paragraph{Emergence of verb-invariant encoding at 14B.}
The $1.000$ subject-control probe ceiling could reflect genuine encoding or verb-identity leakage. A verb-out cross-validation control (Appendix~\ref{sec:appendix_lexical}) holds out each matrix verb and re-trains the probe on the remaining items. At 14B the subject-control probe ceiling generalizes cleanly across held-out verbs. The probe reads the control relation itself rather than memorizing verb identity. The six smaller models show probe-peak gaps deep in the leakage-dominant range. Their ceilings depend on verb identity. Object-control passes verb-out cleanly in every model. The three-level ordering still holds, but the content of its highest level shifts with scale on subject-control. At small scale the highest level is a verb-specific surface readout. At 14B it is a verb-invariant relational encoding. The ``encoded but not decoded'' interpretation of the gap is clean for subject-control at 14B and diluted below it by lexical leakage.

\paragraph{The deployment gap as a layer shift.}
Activation patching across layers (Appendix~\ref{sec:appendix_steering}) locates the gap more precisely than the shortcut account alone. In base models, patching restores controller assignment most effectively at the layer where the probe reads it out. In instruction-tuned models, that effective layer sits ten layers later. Linear steering along the probe-identified direction has no effect, while patching at the shifted layer restores QA subject-control by $+0.37$. Controller information survives instruction tuning as a linear, probe-readable direction. What moves is the layer at which the output can use it. This account accords with mechanistic studies of instruction tuning. Distributional shifts under alignment concentrate in surface-level outputs \citep{lin-etal-2024-unlocking}, and fine-tuning enhances rather than rewires existing mechanisms \citep{prakash-etal-2024-fine}. Whether the surplus compression observed in the other pairs (Results~\S\ref{sec:rq3_instruction}) arises through the same layer shift remains untested.

\paragraph{Methodological implication for syntactic evaluation.}
Behavior-only and probe-only evaluation, often framed as separate windows on form-meaning relations \citep{bender-koller-2020-climbing}, measure different interfaces to whatever the model encodes. LM-head readout is the missing third measurement, turning the behavior--probe gap into a localized diagnosis. The heterogeneous effects of instruction tuning across model pairs (Results~\S\ref{sec:rq3_instruction}) become visible only when the three levels are measured separately. Behavior-only or probe-only evaluation would collapse them into a single score.

\section{Conclusion}
Syntactic evaluation in language models should not equate behavioral failure with the absence of structural information. Encoding, readout, and deployment diverge in a stable order across all seven models and all three languages.

We propose a three-level evaluation framework, which employs behavioral deployment, LM-head readout, and probe recoverability on identical binary decisions, addressing the gap between what a model encodes and what it deploys. These results confirm that the information is in the representation, yet the gap opens at the output. The three-level framework shows that instruction tuning can reshape deployment, not encoding.

\section*{Limitations}
\paragraph{Benchmark scope.}
The benchmark is 48 hand-curated items across English, Chinese, and German, evenly split between object-control and subject-control. It is designed for targeted diagnosis. Template-generated syntax benchmarks such as BLiMP maximize coverage with tens of thousands of items \citep{warstadt-etal-2020-blimp}; this set is the deliberate trade-off, because it holds item identity fixed across the three measurement levels and three languages.

Raw cross-language differences partly reflect tokenization and pretraining exposure rather than typological variation. EUD's explicit \texttt{nsubj:xsubj} propagation is annotated for English but not for Chinese or German, so the controller relation in non-English items is lexically determined by matrix-verb semantics. Cross-condition claims rest on more than the 48 items. The layer-sweep analyses (Appendix~\ref{sec:appendix_steering}) span over $10{,}000$ patched (model, layer, item) observations, and the structural contrast (subject-control gap larger than object-control gap) replicates independently across 7 models, 3 languages, and 4 (task, control-type) cells.

\paragraph{Model coverage.}
The mechanistic evidence is concentrated in the Qwen3 family, where the 0.6B base/instruct pair shows the effect most sharply and Qwen3-14B Instruct rules out a capacity-based interpretation. Llama-3.2 replicates the ordering in an independent family with the complementary degradation profile (\S\ref{sec:rq1_crossfamily}). Cross-family evidence rests on two text families, with Gemma-4 contributing an architectural contrast rather than a third replication; we do not generalize the framework to that architecture class without further evidence. The scaling extensions (Qwen3-1.7B, Llama-3.2-3B) contribute surplus-compression evidence (Results~\S\ref{sec:rq3_instruction}) but no mechanistic data.

\paragraph{Methodological choices.}
Probe recoverability uses linear probes. Nonlinear probes might raise the ceiling without changing the gap direction. LM-head readout assumes the output projection as a fixed map, clean for tied-embedding models and less so otherwise. The counterbalanced-ordering ablation addresses option-position bias but not other prompt-sensitivity effects. Activation patching (Appendix~\ref{sec:appendix_steering}) uses single-layer patches, so we report layer-localization rather than head-level attribution.

\paragraph{Probe ceiling vs. competence.}
The probe summary is an upper bound on what a linear decoder can extract from the most informative layer. It is not a measure of syntactic competence in any full sense, and a high probe score should not be read as evidence that the model would deploy that information under all conditions. The three-level framework is useful precisely because the probe ceiling bounds the other two measurements from above and makes their shortfall visible.

\paragraph{Reproducibility.}
\looseness=-1
All experiments use publicly available pretrained models and ran in fp16 on a single NVIDIA V100 32GB GPU; other GPU architectures may show small numerical drift on near-tie items. The full suite amounts to roughly 25 GPU-hours and 100 CPU core-hours. The 48-item trilingual benchmark, verb-out CV protocol, seeds, and per-language breakdowns are documented in Appendices~\ref{sec:appendix_benchmark}--\ref{sec:appendix_lexical}; probing hyperparameters and the item-out protocol appear in \S\ref{sec:methods}, and activation patching follows the layer-sweep protocol of Appendix~\ref{sec:appendix_steering}.

\nolinenumbers
\let\linenumbers\relax
\bibliography{refs/custom}

\clearpage
\appendix

\section{Descriptive Metric Definitions}
\label{sec:appendix_metrics}

Let $\overline{\mathrm{Acc_B}} = \tfrac{1}{2}\bigl(\mathrm{Acc_B}(\mathrm{QA}) + \mathrm{Acc_B}(\mathrm{Para})\bigr)$ denote mean behavioral accuracy across the two task formats. The recoverability surplus and the behavior-to-recoverability ratio are defined as
\begin{align}
\mathrm{Surplus} &= \mathrm{Acc_P^*} - \overline{\mathrm{Acc_B}}, \\
\mathrm{B/R\ Ratio} &= \overline{\mathrm{Acc_B}} \,/\, \mathrm{Acc_P^*}.
\end{align}

Mean behavior summarizes both task formats rather than selecting the better or worse one, so a single fixed summary avoids choosing a task post hoc. The behavior-to-recoverability ratio expresses how much of the recoverable information reaches the behavioral interface. These quantities are descriptive rather than inferential. They summarize but do not replace the per-task and per-type accuracies in the main tables.

For type-level analyses, let $\overline{\mathrm{Acc_B}}^{\,\mathrm{obj}}$ and $\overline{\mathrm{Acc_B}}^{\,\mathrm{subj}}$ denote mean behavior on the 24 object-control and 24 subject-control items respectively. We define:
\begin{equation}
\mathrm{Subject\text{-}Control\ Deficit} = \overline{\mathrm{Acc_B}}^{\,\mathrm{obj}} - \overline{\mathrm{Acc_B}}^{\,\mathrm{subj}}.
\end{equation}
Positive values mean the model is behaviorally weaker on subject-control than on object-control, independently of its overall deployment level.

\section{Benchmark Examples and Prompt Templates}
\label{sec:appendix_benchmark}

English marks object control with \emph{told} (\emph{John \textbf{told} Mary to leave}; controller \emph{Mary}) and subject control with \emph{promised} (\emph{John \textbf{promised} Mary to leave}; controller \emph{John}). Chinese marks the same contrast with \zh{\cjkbf{让}} (\zh{张三\cjkbf{让}李四离开}; controller \zh{李四}) and \zh{\cjkbf{答应}} (\zh{张三\cjkbf{答应}李四离开}; controller \zh{张三}). German uses \emph{bat} (\emph{Johann \textbf{bat} Maria zu gehen}; controller \emph{Maria}) and \emph{versprach} (\emph{Johann \textbf{versprach} Maria zu gehen}; controller \emph{Johann}).

\begin{table*}[!t]
\centering
\caption{The two behavioral evaluation formats, both using pairwise log-probability scoring over the same candidate controller nouns and differing only in prompt structure.}
\label{tab:task_format}
\small
\setlength{\tabcolsep}{3pt}
\resizebox{\linewidth}{!}{%
\begin{tabular}{p{0.46\textwidth} p{0.46\textwidth}}
\toprule
\textbf{QA Format} & \textbf{Paraphrase Format} \\
\midrule
\textit{Sentence:} ``John promised Mary to leave early.'' &
\textit{Sentence:} ``John promised Mary to leave early.'' \\[4pt]
\textit{Prompt:} \newline
Q: Who is understood to leave early? \newline
A: John $\leftarrow$ \underline{log-prob scored} \newline
A: Mary $\leftarrow$ \underline{log-prob scored} &
\textit{Prompt:} \newline
Which sentence better describes the situation? \newline
(A) John is understood to leave early. $\leftarrow$ \underline{scored} \newline
(B) Mary is understood to leave early. $\leftarrow$ \underline{scored} \\[4pt]
\textit{Decision:} $\arg\max \log P(\text{continuation} \mid \text{sentence} + \text{Q})$ &
\textit{Decision:} $\arg\max \log P(\text{interpretation} \mid \text{sentence} + \text{prompt})$ \\
\bottomrule
\multicolumn{2}{l}{\textit{Gold controller: John} (\emph{promise} = subject-control)} \\
\end{tabular}%
}
\end{table*}

\section{Probe Feature Mode Comparison}

\begin{table}[H]
\centering
\caption{Best per-mode probe accuracy across layers, mean over three seeds.}
\label{tab:appendix_probe_modes}
\small
\begin{tabular}{lccc}
\hline
Mode & \makecell{Qwen3-0.6B\\Base} & \makecell{Qwen3-0.6B\\Instruct} & \makecell{Qwen3-\\14B\\Instruct} \\
\hline
\texttt{ab} & \textbf{0.924} & \textbf{0.847} & 0.958 \\
\texttt{bv} & 0.889 & 0.792 & 0.979 \\
\texttt{av} & 0.618 & 0.458 & 0.931 \\
\texttt{abv} & 0.847 & 0.701 & 0.944 \\
\texttt{v} & 0.757 & 0.521 & \textbf{0.986} \\
\hline
\end{tabular}
\end{table}

Across layers, the choice of feature mode matters more at 0.6B than at 14B, where every mode exceeds $0.93$. The default \texttt{bv} (candidate~B + predicate) stays within $0.06$ of the strongest mode in every model, while \texttt{av} (candidate~A + predicate) is weakest for the small models, consistent with subject-control being harder to encode at the matrix-subject position.

\section{Probe Seed Robustness}

\begin{table}[H]
\centering
\caption{Probe accuracy across three training seeds at the globally-best (layer, feature mode) selected by three-seed mean.}
\label{tab:probe_seed_robustness}
\resizebox{\linewidth}{!}{%
\begin{tabular}{lccccc}
\hline
Model & Seed 1729 & Seed 2718 & Seed 3141 & Mean & Std \\
\hline
Qwen3-0.6B Base & 0.938 & 0.938 & 0.896 & 0.924 & 0.020 \\
Qwen3-0.6B Instruct & 0.833 & 0.812 & 0.896 & 0.847 & 0.035 \\
Qwen3-14B Instruct & 1.000 & 0.979 & 0.979 & 0.986 & 0.010 \\
Llama-3.2-1B Base & 0.812 & 0.833 & 0.833 & 0.826 & 0.010 \\
Llama-3.2-1B Instruct & 0.625 & 0.625 & 0.625 & 0.625 & 0.000 \\
Gemma-4 E4B Base & 0.604 & 0.479 & 0.625 & 0.569 & 0.064 \\
Gemma-4 E4B Instruct & 0.646 & 0.604 & 0.729 & 0.660 & 0.052 \\
\hline
\end{tabular}%
}
\end{table}

Table~\ref{tab:probe_seed_robustness} reports probe accuracy across three training seeds (1729, 2718, 3141) at the globally-best (layer, feature mode) selected by three-seed mean. The standard deviation is over the three seeds at that pick. The main table reports seed 1729 (see Methods~\S\ref{sec:levels}); when the mean-selected pick differs from a seed's own best (layer, feature mode), the seed-1729 value here need not equal the best-probe column of Table~\ref{tab:main_model_comparison}, which is that seed's own best.

\section{Per-Language Breakdown}
\label{sec:appendix_perlang}

\begin{table}[H]
\centering
\caption{Per-language breakdown for all seven models.}
\label{tab:per_language_breakdown}
\resizebox{\linewidth}{!}{%
\begin{tabular}{llcccc}
\hline
Model & Language & Behavior & LM-head & Probe & Surplus \\
\hline
\multirow{3}{*}{Qwen3-0.6B Base} & English & 0.656 & 0.875 & 0.938 & +0.281 \\
 & Chinese & 0.656 & 0.625 & 1.000 & +0.344 \\
 & German & 0.469 & 0.625 & 0.875 & +0.406 \\
\hline
\multirow{3}{*}{Qwen3-0.6B Instruct} & English & 0.688 & 0.875 & 0.812 & +0.125 \\
 & Chinese & 0.469 & 0.562 & 1.000 & +0.531 \\
 & German & 0.375 & 0.562 & 0.688 & +0.312 \\
\hline
\multirow{3}{*}{Qwen3-14B Instruct} & English & 0.719 & 0.688 & 1.000 & +0.281 \\
 & Chinese & 0.844 & 0.875 & 1.000 & +0.156 \\
 & German & 0.812 & 0.812 & 1.000 & +0.188 \\
\hline
\multirow{3}{*}{Llama-3.2-1B Base} & English & 0.625 & 0.812 & 0.812 & +0.188 \\
 & Chinese & 0.719 & 0.812 & 0.938 & +0.219 \\
 & German & 0.375 & 0.438 & 0.688 & +0.312 \\
\hline
\multirow{3}{*}{Llama-3.2-1B Instruct} & English & 0.531 & 0.688 & 0.750 & +0.219 \\
 & Chinese & 0.562 & 0.688 & 0.688 & +0.125 \\
 & German & 0.438 & 0.375 & 0.438 & +0.000 \\
\hline
\multirow{3}{*}{Gemma-4 E4B Base} & English & 0.562 & 0.688 & 0.625 & +0.062 \\
 & Chinese & 0.531 & 0.625 & 0.562 & +0.031 \\
 & German & 0.594 & 0.500 & 0.625 & +0.031 \\
\hline
\multirow{3}{*}{Gemma-4 E4B Instruct} & English & 0.469 & 0.625 & 0.625 & +0.156 \\
 & Chinese & 0.625 & 0.562 & 0.562 & -0.062 \\
 & German & 0.312 & 0.562 & 0.812 & +0.500 \\
\hline
\end{tabular}%
}
\end{table}

Each language contributes 16 items, so a single item swings the per-language accuracy by $1/16 \approx 0.062$. Reading Table~\ref{tab:per_language_breakdown} at this granularity, the three-level ordering $\textit{behavior} \leq \textit{LM-head} \leq \textit{probe}$ holds in every per-language row for the five text-only Qwen3 and Llama-3.2 models up to one-item noise. The only inversions exceeding 0.062 are confined to Gemma-4 E4B (German behavior $0.594$ vs.\ LM-head $0.500$, a 1.5-item swing on a model whose aggregate surplus is already near zero). This is consistent with the boundary interpretation of Gemma-4 in Section~\ref{sec:results} and the Limitations.

The recoverability surplus is not uniform across languages. For the focal Qwen3-0.6B Instruct, surplus on Chinese reaches $+0.531$ while English on the same model is $+0.125$, indicating that the deployment failure is concentrated in non-English data. The same direction holds for the Llama-3.2 base model, whose German surplus is $+0.312$ against an English surplus of $+0.188$. For Qwen3-14B Instruct, German is the cleanest case in which behavior already matches LM-head readout ($0.812$) while the probe still ceilings at $1.000$, isolating a localized deployment shortfall in a high-performing model.

\section{Readout-Cleaning Order Sensitivity}

\begin{table*}[!tb]
\centering
\caption{Readout-cleaning results after counterbalancing option order and label order.}
\label{tab:readout_cleaning}
\resizebox{\linewidth}{!}{%
\begin{tabular}{lcccccc}
\hline
Model & Variant & Summary & Accuracy & Object & Subject & Order Gap \\
\hline
Qwen3-0.6B Base & debiased\_forced\_choice\_ab & debiased & 0.531 & 0.604 & 0.458 & 0.062 \\
Qwen3-0.6B Base & debiased\_label\_choice & debiased & 0.552 & 0.542 & 0.562 & 0.021 \\
Qwen3-0.6B Base & contrastive\_scoring & single & 0.479 & 0.667 & 0.292 & nan \\
Qwen3-0.6B Instruct & debiased\_forced\_choice\_ab & debiased & 0.500 & 0.500 & 0.500 & 0.000 \\
Qwen3-0.6B Instruct & debiased\_label\_choice & debiased & 0.562 & 0.479 & 0.646 & 0.083 \\
Qwen3-0.6B Instruct & contrastive\_scoring & single & 0.500 & 0.708 & 0.292 & nan \\
Qwen3-14B Instruct & debiased\_forced\_choice\_ab & debiased & 0.917 & 0.896 & 0.938 & 0.042 \\
Qwen3-14B Instruct & debiased\_label\_choice & debiased & 0.938 & 0.917 & 0.958 & 0.042 \\
Qwen3-14B Instruct & contrastive\_scoring & single & 0.708 & 0.917 & 0.500 & nan \\
\hline
\end{tabular}%
}
\end{table*}

The debiased values are behavioral accuracies under averaged option order and are not comparable to the best-LM-head column of Table~\ref{tab:main_model_comparison}.

\begin{figure}[t]
    \centering
    \includegraphics[width=\linewidth]{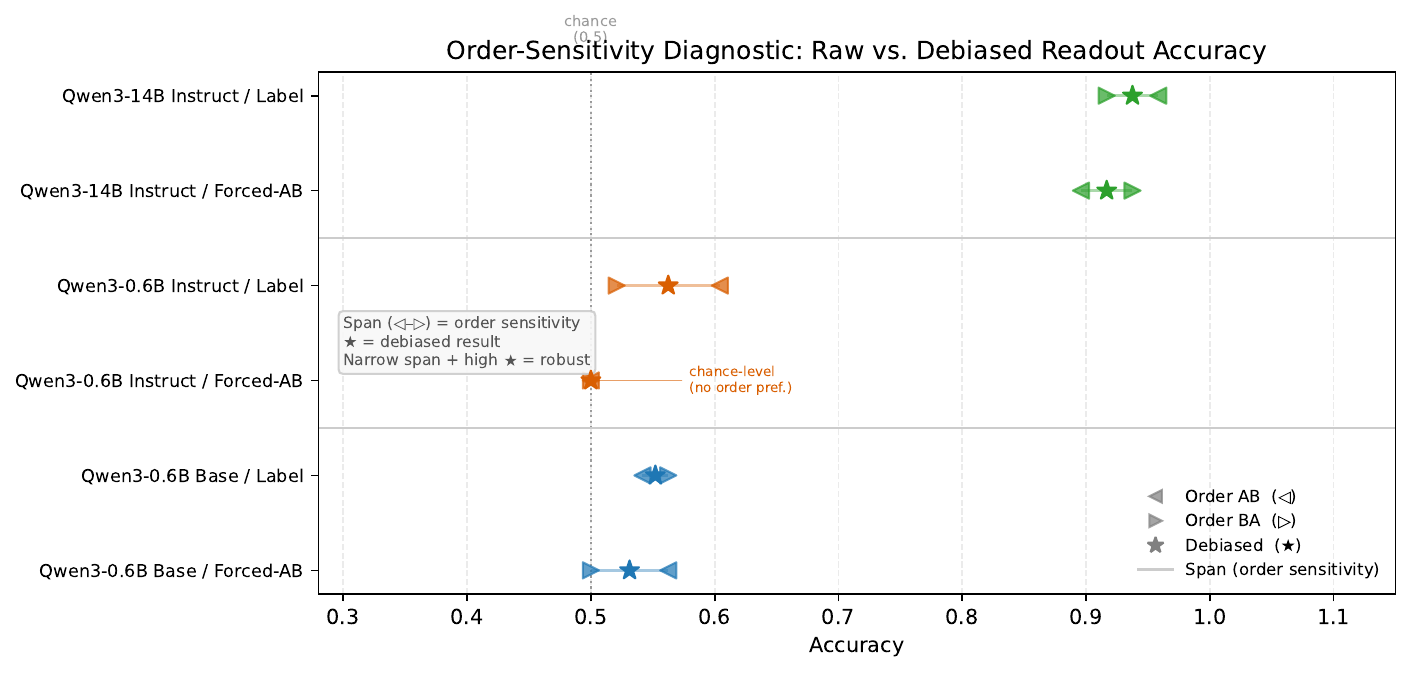}
    \caption{Raw accuracy under both option orders and the debiased result, per model.}
    \label{fig:appendix_cleaning}
\end{figure}

\section{Causal Probing of the Layer-$L$ Geometry}
\label{sec:appendix_steering}

\subsection{Motivation and setup}

The behavior--probe gap admits two interpretations. The LM head fails to deploy controller information that is geometrically present at the probed layer, or the probe identifies geometry that correlates with controller identity but is not the geometry the LM head actually uses. We test both interpretations with two interventions on the residual stream: directional steering (additive perturbation along a probe-derived direction, the amnesic-probing logic of removing probed information and checking the behavioral consequence \citep{elazar-etal-2021-amnesic}) and activation patching \citep{vig-etal-2020-causal,meng-etal-2022-locating} (replacement of one item's hidden state with another's). Directional steering produces no behavioral change at any tested magnitude, while activation patching reveals a layer-localized causal effect that differs systematically between base and instruction-tuned models.

\subsection{Null directional steering at all tested magnitudes}

For each model we compute a steering direction at the layer and feature mode reported as the probe summary in Table~\ref{tab:main_model_comparison}, using two extraction methods: probe weights (logreg) and class-mean difference (diff-of-means, robust to overfitting at $n_{\text{features}} \gg n_{\text{items}}$). The direction is split by mode (e.g., \texttt{ab}: $w_a$ at candidate-A, $w_b$ at candidate-B) and added at each implicated position $p$ as $\alpha \cdot \mathrm{gold\_sign} \cdot \sigma_p \cdot \hat d_p$, where $\mathrm{gold\_sign} \in \{+1, -1\}$ orients $\alpha > 0$ ``toward gold'' for every item. At $\alpha = 0$ the hook reproduces the baseline behavioral scores exactly, validating the implementation.

We swept seven steering strengths from $\alpha = -2$ to $\alpha = +2$ on Qwen3-0.6B Instruct and extended to $\alpha = \pm 3$ and $\pm 5$ to test larger interventions. The $Q_1$ candidate set (subject-control items wrong at $\alpha=0$, $n=18$ on QA) and the $Q_2$ candidate set (object-control items correct at $\alpha=0$, $n=14$) showed no flip at any $\alpha$ for either method. McNemar tests at $\alpha=\pm 5$ versus $\alpha=0$ returned $p=1.000$. Beyond $|\alpha|\geq 3$ a position-bias artifact dominates. Every cell's margin is pushed toward whichever option is favored under heavy off-manifold perturbation. Linear additive steering is therefore not causally sufficient at any tested magnitude.

\subsection{Layer-localized effects of activation patching}

For each (target, donor) pair sharing language and control type, we capture the donor's hidden state at layer $L$ (mode \texttt{abv}, all three positions) and patch it into the target's QA or Paraphrase forward pass, scoring continuations as in Section~\ref{sec:methods}. The causal-rescue lift is the correct-donor rescue rate minus the wrong-donor ``rescue'' rate, isolating donor-correctness from any donor-independent perturbation effect.

\begin{figure*}[!t]
    \centering
    \includegraphics[width=0.92\linewidth]{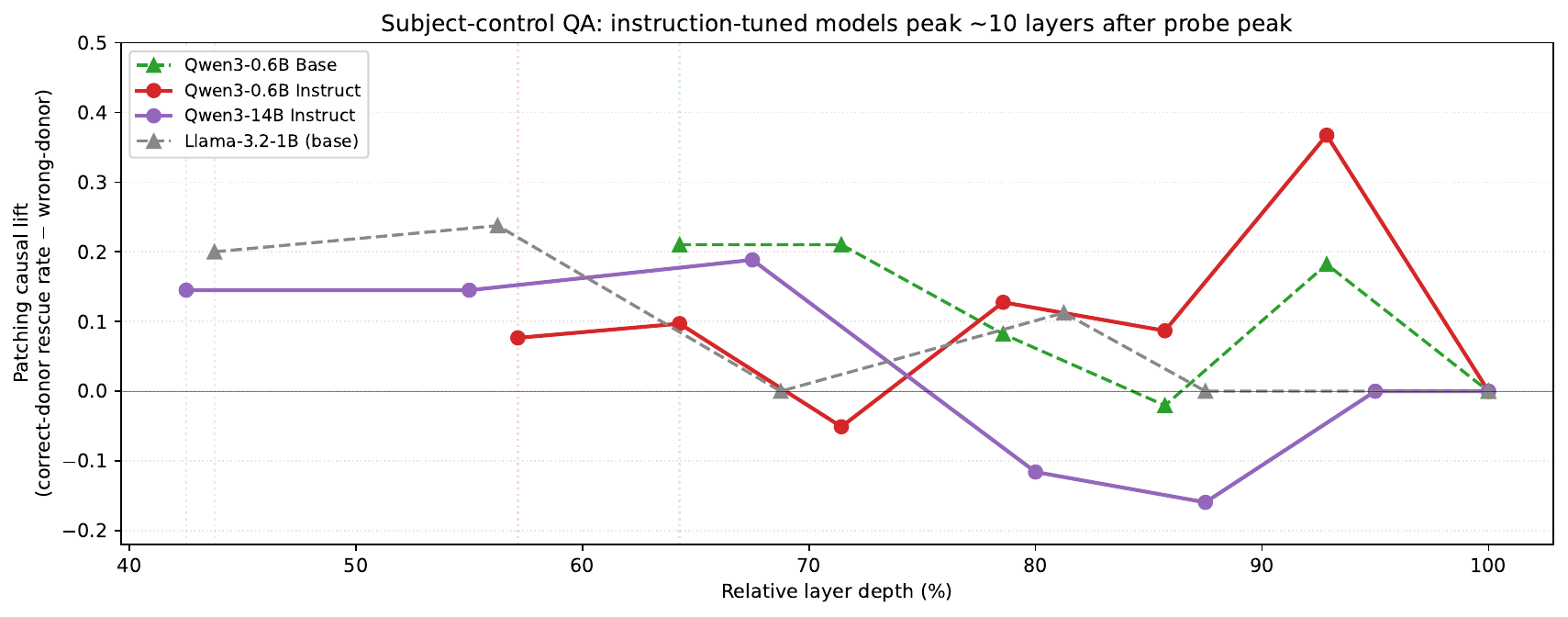}
    \caption{Patching causal lift versus relative layer depth on QA subject-control (vertical dotted lines mark probe peaks).}
    \label{fig:patching_layer_sweep}
\end{figure*}

\begin{figure*}[!t]
    \centering
    \includegraphics[width=0.95\linewidth]{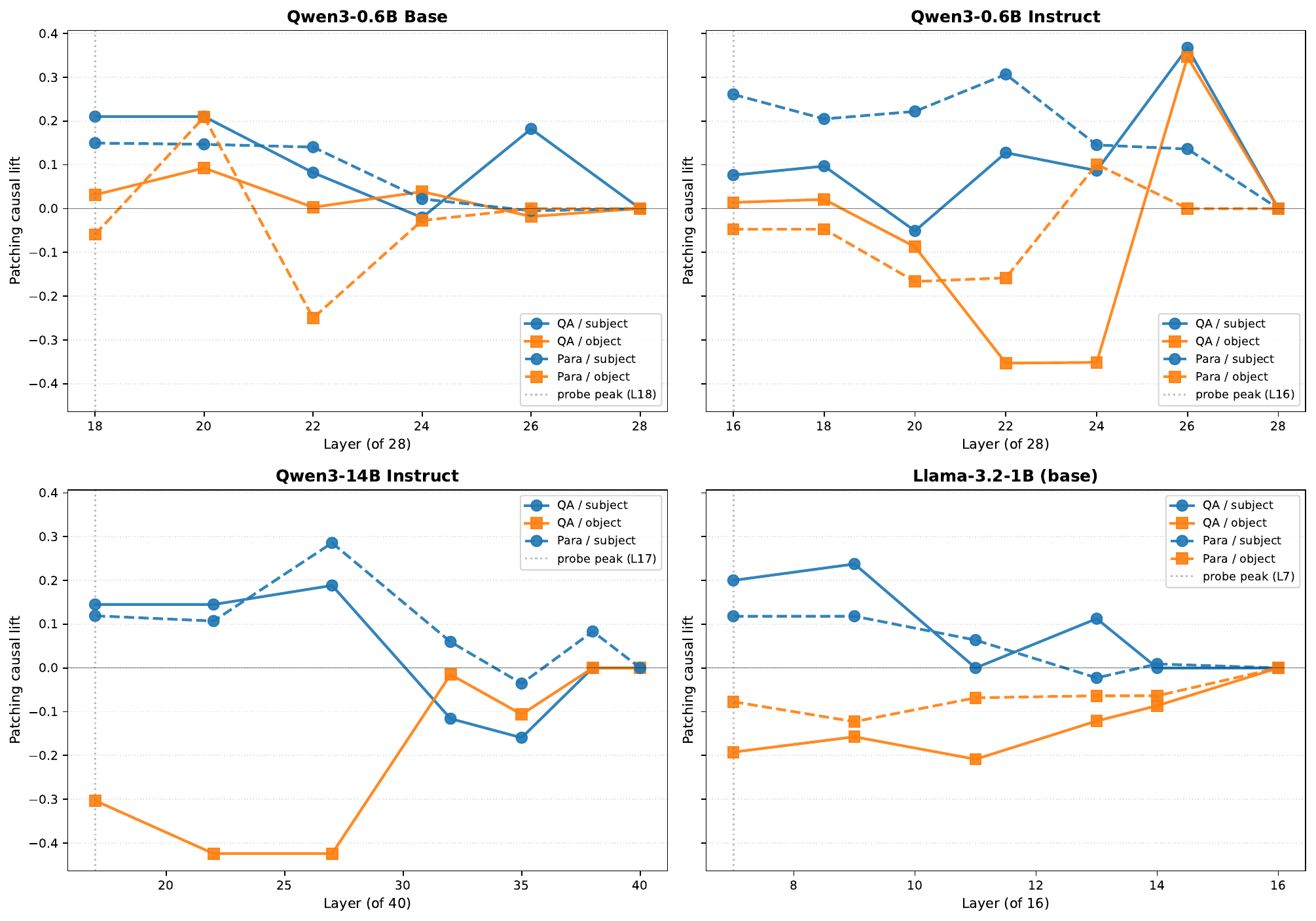}
    \caption{Patching causal lift across all four cells per model.}
    \label{fig:patching_4panel}
\end{figure*}

We swept the patching layer from each model's probe peak through to its final transformer block on four models: two base/no-IT models (Llama-3.2-1B, Qwen3-0.6B Base) and two instruction-tuned ones (Qwen3-0.6B Instruct, Qwen3-14B Instruct). Figure~\ref{fig:patching_layer_sweep} shows the resulting QA subject-control trajectories on a normalized depth axis. Figure~\ref{fig:patching_4panel} shows the per-model breakdown.

\textbf{Result 1.} Llama-3.2-1B base and Qwen3-0.6B Base both show their largest QA subject-control lift at or immediately adjacent to the probe peak (Llama: $+0.24$ at L9, probe at L7; Qwen3-0.6B Base: $+0.21$ at L18--L20, probe at L18) and decay monotonically thereafter. The probe-decoded geometry is, in these models, already the geometry the LM head reads.

\textbf{Result 2.} Qwen3-0.6B Instruct peaks at L26 (probe at L16), with a middle-layer valley at L20 ($-0.05$) and a sharp recovery at L26 ($+0.37$). Qwen3-14B Instruct replicates the qualitative pattern, peaking at L27 (probe at L17), with values around $+0.14$ to $+0.19$ between L22--L27 and declining at L32 and beyond. Both instruction-tuned models peak \emph{ten layers later} than their probe peak.

\textbf{Result 3.} On QA object-control, the two base models show small lifts hovering near zero, whereas Qwen3-0.6B Instruct shows lifts of $-0.35$ at L22--L24 before recovering to $+0.35$ at L26, and Qwen3-14B Instruct shows $-0.42$ at L22 and L27. These negative middle-layer lifts (correct-donor patches do \emph{worse} than wrong-donor patches) point to a regime in which the instruction-tuned model has not yet committed to a controller assignment, where a foreign correct geometry interferes more than a foreign wrong geometry does. The effect is absent in the base models.

\subsection{Joint interpretation of the layer shift}

Linear additive steering is null because the probe-recoverable direction is not the direction the LM head decodes from. Activation patching peaks at the probe layer in base models and ten layers later in instruction-tuned models, identifying a discrete shift in \emph{which layer carries the LM-head-decoded form of the controller geometry}. Taken together, the residual stream encodes controller information at a layer decodable both by a probe and by the LM head in base models. Instruction tuning displaces the LM-head-decoded form to later layers, leaving an intermediate window in which the information is recoverable by probing but not yet usable by output generation. The behavior--probe gap quantified in Section~\ref{sec:results} is the behavioral cost of this displacement. Our results provide a layer-localized empirical instance of the long-standing concern that probe-recoverable directions need not be causally identical to the directions the LM head decodes from \citep{belinkov-2022-probing,hewitt-liang-2019-designing}, and they extend that concern from a general methodological warning into a concrete, controllable layer-shift phenomenon tied to instruction tuning.

The patching lifts at the late peak ($+0.37$ at Qwen3-0.6B Instruct L26 QA subject; $+0.19$ at Qwen3-14B Instruct L27 QA subject) are substantial but below ceiling. Framed in terms of causal abstraction \citep{geiger-etal-2021-causal}, the late-peak layer realizes part but not all of the causal variable corresponding to the controller assignment, so hidden-state geometry at the late peak is causally relevant but not the only causal carrier. Multi-layer simultaneous patching, finer attribution to attention heads or MLP outputs, and a denser layer grid around the late peak remain as immediate causal extensions. The present results establish layer-localization of the deployment gap and its instruction-tuning dependence.

\section{Lexical Leakage Control via Verb-Out Cross-Validation}
\label{sec:appendix_lexical}

\begin{table*}[!tb]
\centering
\caption{Type-level behavior-versus-probe comparison for the most relevant models.}
\label{tab:type_gap_focus}
\resizebox{0.8\linewidth}{!}{%
\begin{tabular}{lccccccc}
\hline
Model & Task & Type & Behavior & Best Probe & Gap & Probe Layer & Probe Mode \\
\hline
Qwen3-0.6B Base & qa & object-control & 0.375 & 0.917 & 0.542 & 18 & ab \\
Qwen3-0.6B Base & qa & subject-control & 0.708 & 0.931 & 0.222 & 18 & ab \\
Qwen3-0.6B Base & paraphrase & object-control & 0.750 & 0.917 & 0.167 & 18 & ab \\
Qwen3-0.6B Base & paraphrase & subject-control & 0.542 & 0.931 & 0.389 & 18 & ab \\
Qwen3-0.6B Instruct & qa & object-control & 0.583 & 0.847 & 0.264 & 16 & ab \\
Qwen3-0.6B Instruct & qa & subject-control & 0.250 & 0.903 & 0.653 & 19 & bv \\
Qwen3-0.6B Instruct & paraphrase & object-control & 0.708 & 0.847 & 0.139 & 16 & ab \\
Qwen3-0.6B Instruct & paraphrase & subject-control & 0.500 & 0.903 & 0.403 & 19 & bv \\
Qwen3-14B Instruct & qa & object-control & 0.833 & 1.000 & 0.167 & 19 & bv \\
Qwen3-14B Instruct & qa & subject-control & 0.792 & 1.000 & 0.208 & 17 & v \\
Qwen3-14B Instruct & paraphrase & object-control & 0.958 & 1.000 & 0.042 & 19 & bv \\
Qwen3-14B Instruct & paraphrase & subject-control & 0.583 & 1.000 & 0.417 & 17 & v \\
\hline
\end{tabular}%
}
\end{table*}

\subsection{Motivation}

Two things could inflate the Qwen3-14B Instruct subject-control probe ceiling of $1.000$ above what genuine encoding of controller information would produce. First, the matrix predicates of the subject-control set cluster. German uses one verb (\zh{versprach}$\times 8$), English splits between \texttt{promised}$\times 4$ and four singletons, and Chinese splits among three verbs. Second, given this clustering, a probe could in principle achieve $1.000$ by memorizing a verb-identity-to-controller mapping --- without the model itself representing control as a syntactic relation. We test this leakage hypothesis directly with Verb-Out Cross-Validation (VO-CV).

\subsection{Verb-Out CV Protocol}

Each unique matrix predicate $V$ defines one fold:
\begin{align*}
\text{train}_V &= \{e \in \mathcal{E} : p(e) \neq V\}, \\
\text{test}_V  &= \{e \in \mathcal{E} : p(e) = V\},
\end{align*}
where $p(e)$ is the matrix predicate of example $e$ and $\mathcal{E}$ is the full 48-item benchmark. The benchmark contains 32 unique predicates across 48 items.

Two design choices matter. First, training pools mix subject- and object-control examples. The probe must learn a non-trivial decision boundary and predictions are aggregated by control type \emph{after} the probe is fitted. Filtering by control type before training would produce a homogeneous label set that any probe trivially solves at $1.000$ even at layer~0, masking rather than testing the leakage hypothesis. Second, because the German subject-control set uses a single matrix verb, holding out \zh{versprach} removes all eight German subject-control items from training simultaneously, so the probe must classify them purely from English and Chinese subject-control plus all object-control evidence. The protocol thus subsumes a strong cross-language generalization test that single-language verb-out CV cannot provide.

We re-extract hidden states with the same pipeline as the main results and re-train the linear probe at the \texttt{v}-mode probe peak (layer~17 of Qwen3-14B Instruct's 40-layer stack) under five feature modes spanning the axis of verb exposure: from \texttt{ab} (both NP candidates, no verb token --- minimum exposure) through \texttt{bv} (candidate B and verb --- the paper's default) to \texttt{v} (verb token only --- maximum exposure). Three random seeds (1729, 2718, 3141) match the main probing protocol (\S\ref{sec:methods}).

\subsection{Focal results for Qwen3-14B Instruct}

Table~\ref{tab:appendix_lexical_leakage} compares item-out LOOCV (the protocol of the main results) against verb-out CV at the Qwen3-14B Instruct probe peak. We treat $|\textit{gap}| < 0.10$ as no leakage, $0.10 \leq \textit{gap} < 0.30$ as partial leakage, and $\textit{gap} \geq 0.30$ as leakage-dominant.

\begin{table}[H]
\centering
\caption{Item-out LOOCV vs.\ verb-out CV at the Qwen3-14B Instruct \texttt{v}-mode probe peak (layer~17, mean $\pm$ SD across three seeds).}
\label{tab:appendix_lexical_leakage}
\resizebox{\linewidth}{!}{%
\small
\begin{tabular}{lcccc}
\hline
Mode & Type & Item-out & Verb-out & Gap \\
\hline
\texttt{ab}  & subj & $0.819 \pm 0.05$ & $0.681 \pm 0.13$ & $+0.139$ \\
\texttt{bv}  & subj & $0.958 \pm 0.04$ & $0.944 \pm 0.06$ & $+0.014$ \\
\texttt{av}  & subj & $0.972 \pm 0.02$ & $0.889 \pm 0.09$ & $+0.083$ \\
\texttt{abv} & subj & $0.903 \pm 0.05$ & $0.889 \pm 0.10$ & $+0.014$ \\
\texttt{v}   & subj & $1.000 \pm 0.00$ & $0.889 \pm 0.05$ & $+0.111$ \\
\hline
\texttt{ab}  & obj  & $0.806 \pm 0.02$ & $0.792 \pm 0.00$ & $+0.014$ \\
\texttt{bv}  & obj  & $0.958 \pm 0.07$ & $0.986 \pm 0.02$ & $-0.028$ \\
\texttt{av}  & obj  & $0.875 \pm 0.00$ & $0.903 \pm 0.02$ & $-0.028$ \\
\texttt{abv} & obj  & $0.847 \pm 0.05$ & $0.861 \pm 0.09$ & $-0.014$ \\
\texttt{v}   & obj  & $0.972 \pm 0.02$ & $0.903 \pm 0.09$ & $+0.069$ \\
\hline
\end{tabular}}
\end{table}

The probe peak survives the verb-out test. In the paper's default \texttt{bv} mode, the gap is $+0.014$ for subject-control and $-0.028$ for object-control. Holding out a matrix verb produces no measurable accuracy drop on either control type. Even the most stringent \texttt{v}-only mode, which feeds the probe only the verb token's hidden state, sees the headline $1.000$ subject-control accuracy fall to $0.889$ under verb-out CV. That accuracy is well above chance ($0.500$), and the associated gap of $+0.111$ stays in the partial-leakage range, below the leakage-dominant threshold. Every cell of Table~\ref{tab:appendix_lexical_leakage} falls in the no-leakage or partial-leakage range, and every object-control row is clean.

\begin{figure*}[!t]
\centering
\includegraphics[width=0.85\linewidth]{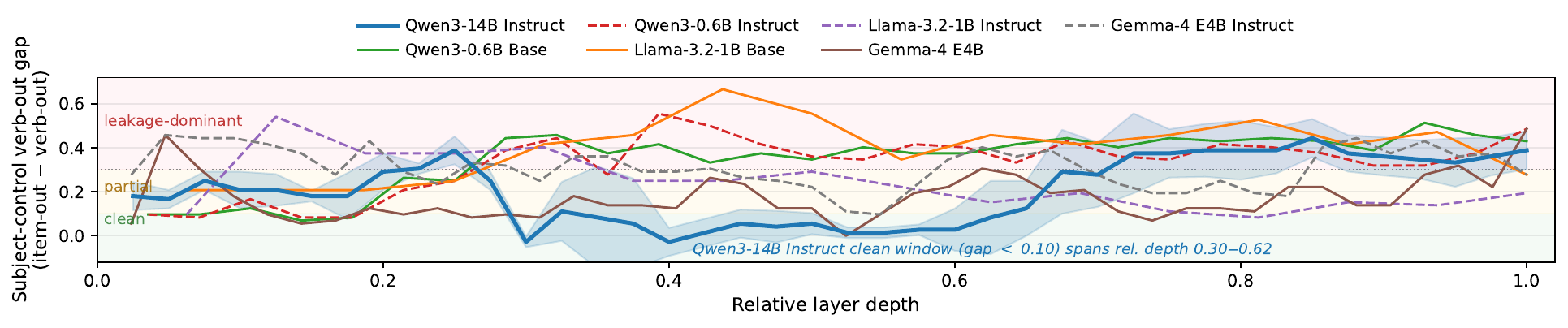}
\caption{Subject-control verb-out gap (item-out LOOCV $-$ verb-out CV) across the full layer trajectory in \texttt{bv} mode.}
\label{fig:appendix_verbout_traj}
\end{figure*}

\subsection{Cross-model verification of scale-dependent emergence}
\label{sec:appendix_lexical_crossmodel}

Across the other six models in our main comparison, the same protocol shows a sharp scale dependence. Table~\ref{tab:appendix_lexical_crossmodel} reports each model's subject-control probe peak in the paper's default \texttt{bv} mode searched across the full layer trajectory along with the verb-out accuracy at the same layer. Object-control is omitted from the table because every object-control probe peak in every model passes the verb-out test cleanly (max object-control gap across the seven models and the full-trajectory layer search is $+0.111$).

\begin{table*}[!t]
\centering
\caption{Subject-control full-trajectory probe peak in \texttt{bv} mode across all seven models, with verb-out CV at the same layer (mean $\pm$ standard deviation across three seeds; gap criterion as in Figure~\ref{fig:appendix_verbout_traj}).}
\label{tab:appendix_lexical_crossmodel}
\resizebox{0.8\linewidth}{!}{%
\begin{tabular}{lccccc}
\hline
Model & Layer & Item-out & Verb-out & Gap & Verdict \\
\hline
Qwen3-0.6B Base       & 18 & $0.917 \pm 0.00$ & $0.500 \pm 0.04$ & $+0.417$ & LEAK \\
Qwen3-0.6B Instruct   & 19 & $0.903 \pm 0.02$ & $0.472 \pm 0.06$ & $+0.431$ & LEAK \\
Llama-3.2-1B Base     &  7 & $0.917 \pm 0.00$ & $0.250 \pm 0.00$ & $+0.667$ & LEAK \\
Llama-3.2-1B Instruct &  2 & $0.625 \pm 0.00$ & $0.083 \pm 0.00$ & $+0.542$ & LEAK \\
Gemma-4 E4B           &  2 & $0.611 \pm 0.06$ & $0.153 \pm 0.02$ & $+0.458$ & LEAK \\
Gemma-4 E4B Instruct  &  4 & $0.750 \pm 0.00$ & $0.306 \pm 0.06$ & $+0.444$ & LEAK \\
\textbf{Qwen3-14B Instruct}    & 22 & $\mathbf{0.986 \pm 0.02}$ & $\mathbf{0.972 \pm 0.02}$ & $\mathbf{+0.014}$ & \textbf{clean} \\
\hline
\end{tabular}%
}
\end{table*}

Object-control passes the verb-out test cleanly at every model and layer. Subject-control does not. Only Qwen3-14B Instruct reaches a layer where high probe accuracy and clean verb-out hold simultaneously. Every smaller model lands in the leakage-dominant range at its full-trajectory probe peak. Three smaller models (Qwen3-0.6B Base, Qwen3-0.6B Instruct, Llama-3.2-1B Base) reach $0.903$--$0.917$ item-out at their probe peaks yet still fail verb-out, so the issue is what the probe extracts rather than raw probe accuracy. Clean verb-out at low item-out, observed at the embedding layer of every model, identifies a probe that has learned nothing. Clean verb-out at high item-out, achieved only by Qwen3-14B Instruct, identifies representational structure that generalizes across held-out matrix verbs.

Figure~\ref{fig:appendix_verbout_traj} shows the full subject-control verb-out gap trajectory across all transformer layers for the seven models. Qwen3-14B Instruct's clean window is not a single layer but a contiguous plateau from relative depth $0.30$ to $0.62$ (layers $12$--$25$ of 40), in which both item-out and verb-out accuracies stay above $0.9$ and their gap stays below the $0.10$ threshold. The six smaller models' trajectories never enter the clean band. Their probe-peak gaps in Table~\ref{tab:appendix_lexical_crossmodel} reflect the lowest point of trajectories that otherwise stay between the partial-leakage and leakage-dominant bands at every depth.

The three-level $\textit{behavior} \leq \textit{LM\text{-}head} \leq \textit{probe}$ ordering holds across all seven models, but the \emph{content} of the highest level is scale-dependent for subject-control: a verb-specific lookup at small scale, a verb-invariant relation at 14B (layer~17 in \texttt{v} mode, layer~22 in \texttt{bv} mode) that generalizes across held-out matrix verbs and language boundaries. Object-control is verb-invariant in every model. The deployment-gap framework holds at every scale. The strong interpretation of the subject-control probe ceiling as model-encoded geometry is clean at 14B and diluted below it by lexical leakage.

\section{Type-Level Gap and Three-Level Framework}
\label{sec:appendix_typegap}

Table~\ref{tab:type_gap_focus} reports the per-(model, task, control-type) behavior and probe accuracy that underlie the type-level analysis of Section~\ref{sec:rq2_subject14b}, with the probe best over layers and feature modes (seed 1729). The Qwen3-14B Instruct subject-control probe ceiling of $1.000$ survives verb-out CV (Appendix~\ref{sec:appendix_lexical}). Figure~\ref{fig:framework} illustrates the three-level separation on the sharpest single case.

\end{document}